\pdfoutput=1
\documentclass[11pt]{article}

\DeclareUnicodeCharacter{03BB}{\ensuremath{\lambda}}

\newif\ifarxiv
\arxivtrue % make it 'true' or 'false' if you want to keep the public or anonymized version

\ifarxiv
\usepackage[preprint]{acl}
\else
\usepackage[review]{acl}
\fi

\usepackage{times}
\usepackage{latexsym}

\usepackage[T1]{fontenc}
\usepackage[utf8]{inputenc}

\usepackage{microtype}

\usepackage{inconsolata}

\usepackage{graphicx}
\usepackage{booktabs}       % professional-quality tables
\usepackage{amsfonts}       % blackboard math symbols
\usepackage{nicefrac}       % compact symbols for 1/2, etc.
\usepackage{microtype}      % microtypography
\usepackage{xcolor}         % colors

\usepackage{graphicx}
\usepackage{amsmath}
\usepackage[inline]{enumitem}
\usepackage[outline]{contour}
\usepackage{multirow}
\usepackage{wrapfig}
\usepackage{subfigure}
\usepackage{diagbox}
\usepackage{enumitem}
\usepackage{xcolor}
\usepackage{tcolorbox}
\usepackage{tabularx}

\usepackage[textsize=tiny]{todonotes}

\newtcolorbox{examplebox}[1][]{
    coltitle=black,
    colback=blue!5,
    colframe=blue!30,
    fonttitle=\bfseries,
    colbacktitle=blue!30,
    title={Prompt},
    #1
}

\definecolor{gred}{HTML}{cc0200}
\definecolor{ggreen}{HTML}{2FD156}  
\newcommand{\ua}[1]{\scriptsize\textcolor{ggreen}{\footnotesize $\uparrow$}{\color{ggreen}#1}} %{\myfontsize $\%$}
\newcommand{\gray}[1]{\textcolor{gray}{#1}}

\title{The Reasoning-Memorization Interplay in Language Models \\ Is Mediated by a Single Direction}

\author{Yihuai Hong$^1$\footnotemark[1] \quad Dian Zhou$^2$ \quad Meng Cao$^3$ \quad Lei Yu$^{5}$\footnotemark[2] \quad Zhijing Jin$^{4,5,6}$\footnotemark[2] \\ [10px]
        $^1$New York University \   $^2$University of Illinois at Urbana-Champaign  \\ 
        $^3$McGill University  \ $^4$Max Planck Institute for Intelligent Systems, Tuebingen, Germany  \\ $^5$University of Toronto \ $^6$Vector Institute \\  [5px]
        }

\begin{document}
\maketitle

\renewcommand{\thefootnote}{\fnsymbol{footnote}}

\footnotetext[1]{Work done prior to joining New York University.}
\footnotetext[2]{Corresponding authors.}

\begin{abstract}
Large language models (LLMs) excel on a variety of reasoning benchmarks, but previous studies suggest they sometimes struggle to generalize to unseen questions, potentially due to over-reliance on memorized training examples. However, the precise conditions under which LLMs switch between reasoning and memorization during text generation remain unclear. In this work, we provide a mechanistic understanding of LLMs’ reasoning-memorization dynamics by identifying a set of linear features in the model's residual stream that govern the balance between genuine reasoning and memory recall. These features not only distinguish reasoning tasks from memory-intensive ones but can also be manipulated to causally influence model performance on reasoning tasks. Additionally, we show that intervening in these reasoning features helps the model more accurately activate the most relevant problem-solving capabilities during answer generation. Our findings offer new insights into the underlying mechanisms of reasoning and memory in LLMs and pave the way for the development of more robust and interpretable generative AI systems.
% \footnote{Our code and data 
% \ifarxiv
% are at \url{https://github.com/causalNLP/}.
% \else
% have been uploaded to the submission system, and will be open-sourced upon acceptance.
% \fi
% }

\end{abstract}
\section{Introduction}

Large language models (LLMs) have demonstrated impressive capabilities in tackling complex reasoning tasks \citep{roziere2023code, openai2024learning, guo2025deepseek}. However, these models sometimes struggle with more straightforward reasoning problems, particularly when faced with questions that differ significantly from those encountered during training \citep{dziri2024faith, hu2024case, xie2024memorizationlargelanguagemodels}. This generalization gap between LLMs and human reasoning has led to the hypothesis that these models are essentially ``reasoning parrots'' \citep{zečević2023causalparrotslargelanguage}, relying heavily on \textit{memorization} of text patterns found in their pretraining datasets \citep{carlini2022quantifying, tang2023large, shi2023detecting}, rather than engaging in a rigorous, procedural reasoning process to solve problems \citep{wei2022chain, kojima2022large, yao2023react}. Understanding the interplay between reasoning and memorization in LLMs is essential, not only for advancing our understanding of these models but also for developing more reliable, language-based reasoning systems in the future \citep{lanham2023measuring, oren2023proving, turpin2024language}.

\begin{figure}[t]
    \centering
    \includegraphics[width=0.5\textwidth]{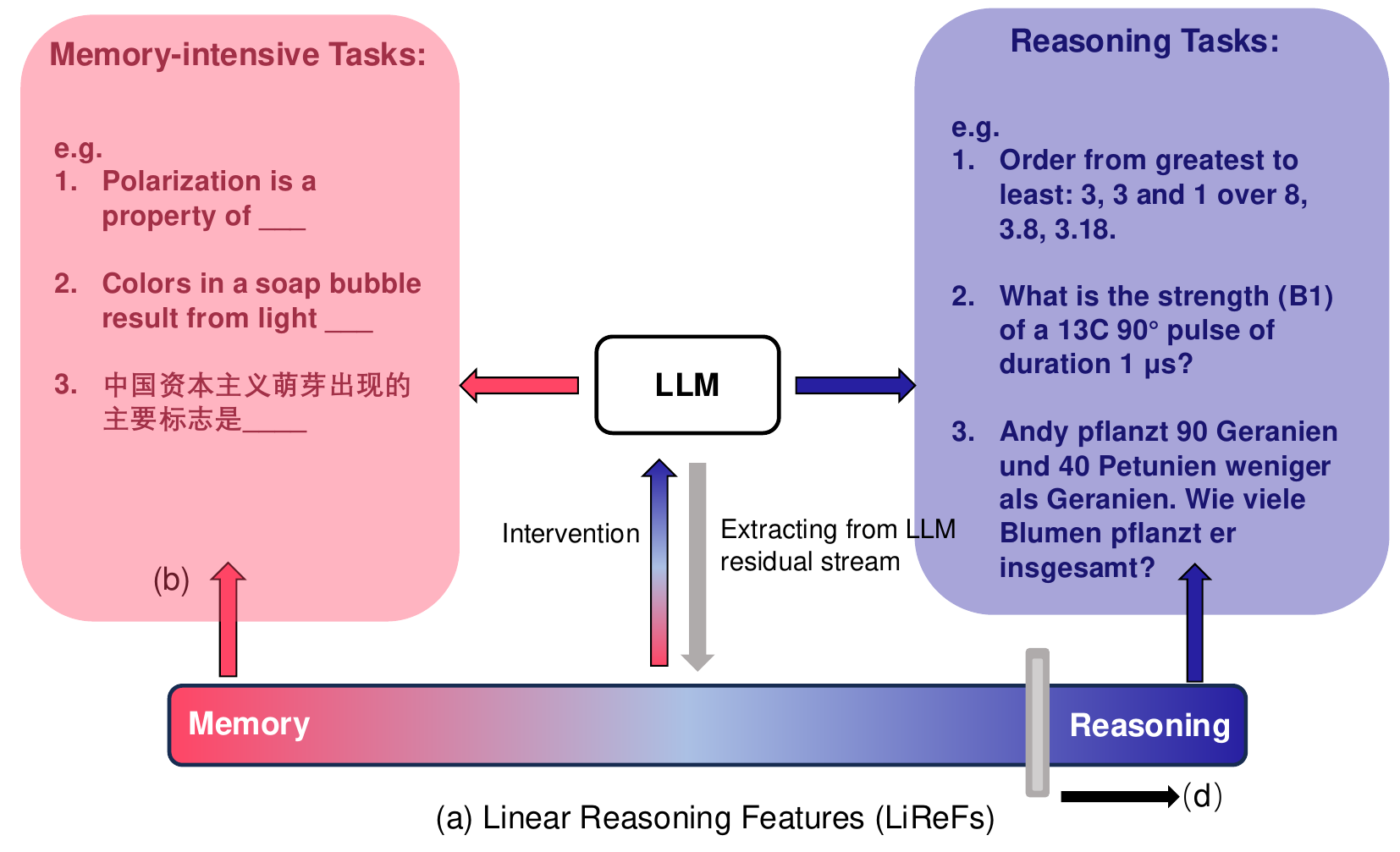}
    \caption{Main findings of our study: (a) There exists a set of linear features (LiReFs) in the LLM residual stream that drives the model to switch between reasoning and memorization modes with different levels of generalizability.
(b) LiReFs generally explain model reasoning capability across various knowledge domains and languages.
(c) Model activation values along LiReFs correlate strongly with model generalizability on reasoning tasks.
(d) Intervening LiReFs during inference time can further improve the model reasoning performance and generalizability.}
    \label{fig:LiReFs}
\end{figure}

In the context of LLM reasoning, researchers often conceptualize memorization as the inability to generalize from familiar problems to their systematically modified counterparts. In this view, reasoning and memorization are two extremes on the spectrum of model generalizability. To investigate this, synthetic reasoning benchmarks are designed, and memorization is assessed by measuring changes in model performance across various setups \citep{dziri2024faith, xie2024memorizationlargelanguagemodels, ye2024physicslanguagemodels21}. Another line of research focuses on the internal mechanisms of LLMs, identifying specific components or circuits responsible for tasks like arithmetic \citep{hou2023towards, stolfo2023mechanistic} and commonsense reasoning \citep{geva-etal-2023-dissecting, yang2024large, biran2024hopping}. However, these studies primarily analyze model outputs or hidden representations when dealing with carefully crafted synthetic reasoning problems, limiting the generalizability of their findings.

In this paper, we explore the reasoning-memorization dynamic of LLMs from a mechanistic perspective. Recent interpretability research has demonstrated that LLMs encode interpretable semantic features \citep{elhage2022toy, park2024linearrepresentationhypothesisgeometry}—such as safety \citep{arditi2024refusal, yu2024robust}, truth \citep{marks2023geometry, li2024inference}, sentiment \citep{tigges2023linear}, and language \citep{bricken2023monosemanticity}—as linear directions within their activation space. We hypothesize that there is a similar linear feature, which, when activated, enables the model to solve reasoning tasks through systematic generalization. When this feature is not activated, the model remains in a ``memorization mode,'' exhibiting low generalizability when addressing variations of familiar reasoning problems.

To examine our hypotheses, we apply methods from linear semantic feature analysis \citep{burnsdiscovering,rimsky-etal-2024-steering} and identify a set of Linear Reasoning Features (LiReFs) in the residual streams of LLMs. As shown in Figure \ref{fig:LiReFs}, LiReFs can be extracted by contrasting the hidden representations of reasoning-intensive versus memory-intensive questions. This contrast allows the two types of questions to be linearly separated in the model's activation space. Furthermore, we demonstrate via causal analysis \citep{stickland2024steering,hong2024intrinsic} that by enhancing the LiReFs during inference, we can shift the model into a ``thinking mode'' with strong generalizability in applying reasoning rules or patterns. We show via extensive experiments on four different LLMs across six datasets that the same set of reasoning features explain and mediate model reasoning ability across various knowledge domains and languages, suggesting a general control mechanism of switching between reasoning and memorization during model inference.

The main contributions of our work can be summarized as follows:

\begin{itemize}
    \item We show that LLM reasoning capability is mediated by a set of linear features (LiReFs) in its activation space. Such features govern model generalizability in solving various reasoning tasks including math, logical, and scientific questions (Section \ref{sec:liref}).
     \item We casually validate the functionality of our discovered reasoning features by showing that LLM reasoning generalizability can be enhanced by intervening LiReFs at inference time (Section \ref{sec: intervention}).
    \item We show via case analyses that mediating LiReFs during inference time reduces LLM reasoning errors and misapplication of model reasoning or memorization ability. (Section \ref{sec: cases analysis}). 
\end{itemize}

% In \S\ref{sec: reason generalization}, by intervening the reasoning features indentified, we can effectively influence the model's performance and generalization on the reasoning dataset. By further directing these representations towards the reasoning region in the semantic space, the model achieves improved performance and stronger generalization on this corresponding reasoning dataset.

\section{Related work}

\paragraph{Memorization in LLMs}
Memorization in LLMs has been defined in various ways. In the context of privacy and copyright, memorization is often described as the model's verbatim reproduction of training data during generation \citep{carlini2022quantifying, biderman2023emergentpredictablememorizationlarge, huang2024demystifyingverbatimmemorizationlarge}. Alternatively, some define memorization as the counterfactual effect of omitting specific training data on model predictions \citep{zhang2023counterfactual, hu2024case}, reflecting memorization of rare, specific examples. In reasoning tasks, memorization is often seen as poor generalizability to questions outside the training data, as evidenced by studies on work sequence reversal \citep{mccoy2023embersautoregressionunderstandinglarge} and alphabet shifting \citep{prabhakar2024decipheringfactorsinfluencingefficacy}, which show degraded performance on infrequent patterns. Other studies observe performance degradation from controlled perturbations of input questions \citep{wu2024reasoningrecitingexploringcapabilities, xie2024memorizationlargelanguagemodels}. In this paper, we adopt memorization as poor reasoning generalizability and propose a novel mechanistic interpretation of the reasoning-memorization dynamic during model inference.

\paragraph{Understanding LLM reasoning} 
% summarize papers in section 2.2 of Yihuai's related work document

% 这里是需要我们提到对reason的定义的，是知识 + 规则，具有泛化性，至少能够超越domain，etc

% \yh{Revision Needed. TODO.}
Prior research has sought to distinguish reasoning from memorization, investigating whether LLMs genuinely infer new conclusions or merely reconstruct patterns from pretraining data. Studies suggest that LLMs undergo structured multi-step reasoning processes, transitioning through distinct reasoning stages that follow an ordered sequence of knowledge retrieval and rule-based processing \citep{hou2023towards}. Similarly, extended training beyond overfitting (grokking) has been shown to lead to the emergence of reasoning circuits, indicating that reasoning is a learned and structured capability \citep{power2022grokking,liu2022towards,nanda2023progress,wang2024grokked}. Further studies on mathematical reasoning confirm that LLMs compute necessary information rather than memorizing templates, with reasoning computations leaving identifiable traces in model activations, particularly in the residual stream \citep{ye2024physicslanguagemodels21, stolfo-etal-2023-mechanistic}. Additionally, attention heads have been shown to play a key role in both knowledge recall and latent reasoning, suggesting that these processes are distinct yet interconnected \citep{zheng2024attentionheadslargelanguage}.

\paragraph{Linear semantic features}
Recent advances in model interpretability have revealed that language models encode various semantic concepts as linear directions in their activation space \citep{park2024linearrepresentationhypothesisgeometry}. These linear semantic features have been discovered by contrasting inputs that differ primarily in the targe semantic dimension \citep{marks2023geometry}. Once identified, these linear features can be manipulated to control model behavior, enabling targeted interventions during the generation process \citep{rimsky-etal-2024-steering,stickland2024steering}. Our work extends this line of study by identifying linear features that mediate the model's ability to switch between genuine reasoning and memory recall. 
\section{Linear reasoning features (LiReFs)}
\label{sec:liref}
\subsection{Background}
\paragraph{Transformers}
A decoder-only transformer language model ~\citep{vaswani2017attention} $\mathcal{M}$ maps an input sequence of tokens $x = [x_1,...,x_T]$ into a probability distribution over the vocabulary for next-token prediction. Within the transformer, the $i$-th token $x_i$ is represented as a series of hidden states $\mathbf{h}^{(l)}(x_i)$. Within each layer $l \in [L]$, two modules compute updates that are added to the layer input $\mathbf{h}^{(l-1)}(x_i)$: (1) a \textbf{multi-head self-attention module} outputs $\mathbf{a}^{(l)}(x_i)$, and a \textbf{multi-layer perceptron (MLP)} outputs $\mathbf{m}^{(l)}(x_i)$. Putting together, the hidden representation $\mathbf{h}^{(l)}(x_i)$ is computed as \footnote{Here, we omit some details such as positional encoding and layer normalization for brevity.}: 
\begin{align}
    \mathbf{h}^{(l)}(x_i) = \mathbf{h}^{(l-1)}(x_i) + \mathbf{a}^{(l)}(x_i) + \mathbf{m}^{(l)}(x_i)
\end{align}
Following \citet{elhage2021mathematical}, we call each $\mathbf{h}^{(l)}(x_i)$ the \textit{residual stream activation} of $x_i$ at layer $l$. We focus on the residual stream of the last token $x_T$ of the user turn, as the point when the model is going to generate the first answer token, denoted as $\mathbf{H}(x)=\{\mathbf{h}^{(l)}(x_T)\}_{l=1}^{L}$.

\paragraph{Reasoning feature extraction}
We follow the linear feature hypothesis and postulate that the reasoning capability of LLMs is mediated by a single direction in the residual stream, and that by steering this direction, it is possible to control model interplay between reasoning and memorization. We compute the \textit{linear reasoning features (LiReFs)} using the \textit{difference-in-means} technique, which effectively disentangles key feature information as demonstrated by previous work \citep{marks2023geometry,rimsky-etal-2024-steering}. Specifically, given a collection of \textit{reasoning-intensive questions} $x \in \mathcal{D}_\textbf{Reasoning}$ (e.g. ``What is the answer of $(5+2)*3$?'') and another set of \textit{memory-intensive questions} $x \in \mathcal{D}_\textbf{Memory}$ (e.g. ``What is the capital city of the USA?''), we calculate the difference between the model’s mean last-token residual stream activations when running on two categories of input questions:
\begin{equation}
\resizebox{0.89\linewidth}{!}{
    $\mathbf{r}^{(l)} = \frac{\sum\limits_{x\in \mathcal{D}_\textbf{Reasoning}} \mathbf{h}^{(l)}(x)}{|\mathcal{D}_\textbf{Reasoning}|} - \frac{\sum\limits_{x\in \mathcal{D}_\textbf{Memory}} \mathbf{h}^{(l)}(x)}{|\mathcal{D}_\textbf{Memory}|}$}
\label{eq:hh-refusal-feature}
\end{equation}
% where we construct $\mathcal{D}_\text{reason}$ and $\mathcal{D}_\text{memory}$ by sampling \_\_ instructions from the \_\_ dataset and the \_\_ dataset respectively. 
The specific construction details of $\mathcal{D}_\textbf{Memory}$ and $\mathcal{D}_\textbf{Reasoning}$ are provided in Section \ref{sec: datasets and models}.

\paragraph{Reasoning feature intervention}
Given a difference-in-means vector $\mathbf{r}^{(l)}$ extracted from layer $l$, we can modulate the strength of the corresponding reasoning feature via simple linear interventions. Specifically, we can perform \textit{reasoning feature addition} by adding the difference-in-means vector to the activations of an input question to shift it closer to the mean activation of typical reasoning-intensive questions, thereby unlocking model reasoning capability:
\begin{align}
    \mathbf{h'}^{(l)}(x) \leftarrow \mathbf{h}^{(l)}(x) + \alpha * \mathbf{r}^{(l)}
\label{eq:feature-addition}
\end{align} 

Similarly, one can perform \textit{reasoning feature ablation} by erasing the component along $\hat{\mathbf{r}}^{(l)}$ for every residual stream activation $\mathbf{h}^{(l)}(x)$:
\begin{align}
    \mathbf{h'}^{(l)}(x) \leftarrow \mathbf{h}^{(l)}(x) - \hat{\mathbf{r}}\hat{\mathbf{r}}^T\mathbf{h}^{(l)}(x)
\label{eq:feature-ablation}
\end{align}
where $\hat{\mathbf{r}} = \mathbf{r}^{(l)} / ||\mathbf{r}^{(l)}||$ is a unit vector encoding the reasoning feature direction, and $\mathbf{h}^{(l)}(x) - \hat{\mathbf{r}}\hat{\mathbf{r}}^T\mathbf{h}^{(l)}(x)$ is projection that zeroes out the value along the reasoning direction.

% step1 模型在两个对抗的数据训练集上run并且取activations (hiddenstates), 相减后取平均, 在每一层上得到all candidate features

% step2 selecting the best features by evaluating each single feature on the validation set

\subsection{Datasets and Models}
\label{sec: datasets and models}

\paragraph{Datasets} We curate our dataset for LiReF extraction and analysis using the following existing question answering benchmarks: 1) MMLU-Pro \citep{wang2024mmlupro}, which is a comprehensive QA benchmark covering a wide range of subjects, including STEM, humanities and social sciences fields; 2) the GSM-8K math reasoning dataset \citep{cobbe2021trainingverifierssolvemath} and its multilingual counterpart MGSM \citep{shi2022languagemodelsmultilingualchainofthought}; 3) the PopQA factual knowledge QA dataset \citep{mallen-etal-2023-trust}, and 4) the humanity sections of the C-Eval Chinese benchmark \citep{huang2023ceval}. A detailed description of each dataset can be found in \S\ref{appendix: details of datasets}.

To categorize QA questions into the contrastive reasoning-intensive and memory-intensive subsets, we employ LLM-as-a-judge \citep{zheng2023judging} by asking GPT-4o \citep{openai2024gpt4ocard} to assign a score between 0 and 1 to each question in MMLU-Pro, where a score closer to 1 indicates a reasoning-intensive question, and a score closer to 0 suggests a memory-intensive one. A score around 0.5 indicates that both reasoning and memory recall may be involved \footnote{The prompt used is provided in \S\ref{appendix: prompt}.}. Next, we classified questions with scores above 0.5 as MMLU-Pro-R (Reasoning Part) and placed them in $\mathcal{D}_\textbf{Reasoning}$, while questions with scores less than or equal to 0.5 were classified as MMLU-Pro-M (Memory Part) and placed in $\mathcal{D}_\textbf{Memory}$. For the other benchmarks, we assign GSM8K and MGSM into $\mathcal{D}_\textbf{Reasoning}$, and put PopQA and C-Eval Chinese into $\mathcal{D}_\textbf{Memory}$.

\paragraph{Models}

We study LiReF by analyzing a diverse collection of representative and influential base models, as long as their instruction-tuned variants: LLaMA3-8B (base, instruct) \citep{grattafiori2024llama3herdmodels}, Gemma2-9B (base, instruct) \citep{gemmateam2024gemma2improvingopen}, Mistrial-7B-v0.3 (base, instruct) \citep{jiang2023mistral7b}, and OLMo2-7B (base, instruct) \citep{olmo20252olmo2furious}.

\subsection{Analysis results}
\label{sec: robustness analysis}

\begin{figure*}[t]
\setlength{\belowcaptionskip}{-8px}
    \centering
    \includegraphics[scale=0.34]{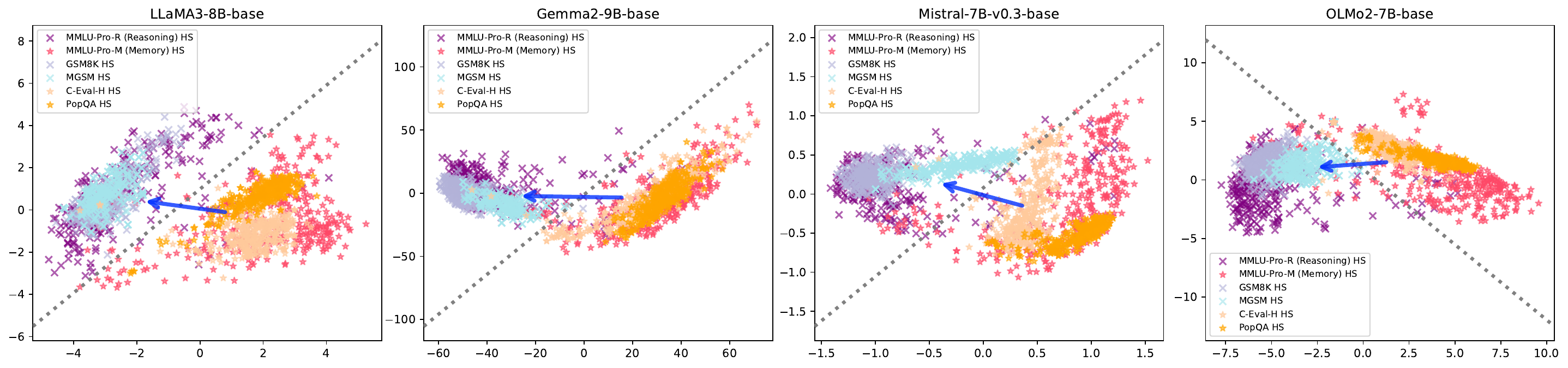}
    \caption{Visualization of the hidden states of four base models using 2-dimensional PCA. For each model, we plot six groups of points across several datasets. We observe that: (1) For all four models, questions defined as Reasoning-required and those defined as Memory-required can be naturally distinguished into two distinct groups, as shown by the boundary (grey dashed line) fitted via logistic regression, with the blue arrows showing the approximate direction of the Linear Reasoning Features. (2) In the extracted dimensions, the influence of task domain and language within the same category on the distribution is not significant, and data requiring the same capability naturally cluster together in the same region.}
    \label{fig: pca_on_reason_memory}
\end{figure*}

Figure \ref{fig: pca_on_reason_memory} shows the 2-dimensional Principal Component Analysis (PCA) visualization of the last tokens representations across different model layers and six datasets in $\mathcal{D}_\textbf{Memory}$ and $\mathcal{D}_\textbf{Reasoning}$, where hidden representations are taken from a specific middle layer of each model. \footnote{Figure \ref{fig: cosine_pca_mean_diff} in the Appendix \ref{appendix: additional experiments} shows that the top one principal component already captures most of the mean difference (see Equation \ref{eq:hh-refusal-feature}) between the activations in $\mathcal{D}_\textbf{Memory}$ and $\mathcal{D}_\textbf{Reasoning}$.} Additional PCA results for other layers of the models are provided in Appendix \ref{appendix: additional experiments}. We observe that the representations of questions in $\mathcal{D}_\textbf{Memory}$ and $\mathcal{D}_\textbf{Reasoning}$ can be linearly separated by the reasoning features, which are computed as the difference vector between centroids of the two representation categories (the blue arrows). 

\paragraph{Robustness of LiReF extraction}
We also validate that our extracted LiReFs indeed capture model reasoning capability, as opposed to some superficial lexical patterns that distinguish two question categories. As suggested by Figure \ref{fig: pca_on_reason_memory}, for each model, the same LiReF separates every contrastive pair of problem subsets in $\mathcal{D}_\textbf{Reasoning}$ and $\mathcal{D}_\textbf{Memory}$, regardless of the task format (e.g., multiple choice and the open-ended generation), domain (e.g., physics, chemistry and math), or language (e.g., English and Chinese). Moreover, we provide in Appendix \ref{appendix: additional experiments} more fine-grained PCA visualizations of questions from various subject domains in MMLU-Pro, suggesting that even for questions from disparate disciplines (e.g., physics vs. history), as long as both of their solutions require strong reasoning capability, their hidden representations shall fall into the same reasoning subspace as determined by the LiReF. 

\begin{figure*}[t]
\setlength{\belowcaptionskip}{-8px}
    \centering
    \includegraphics[scale=0.28]{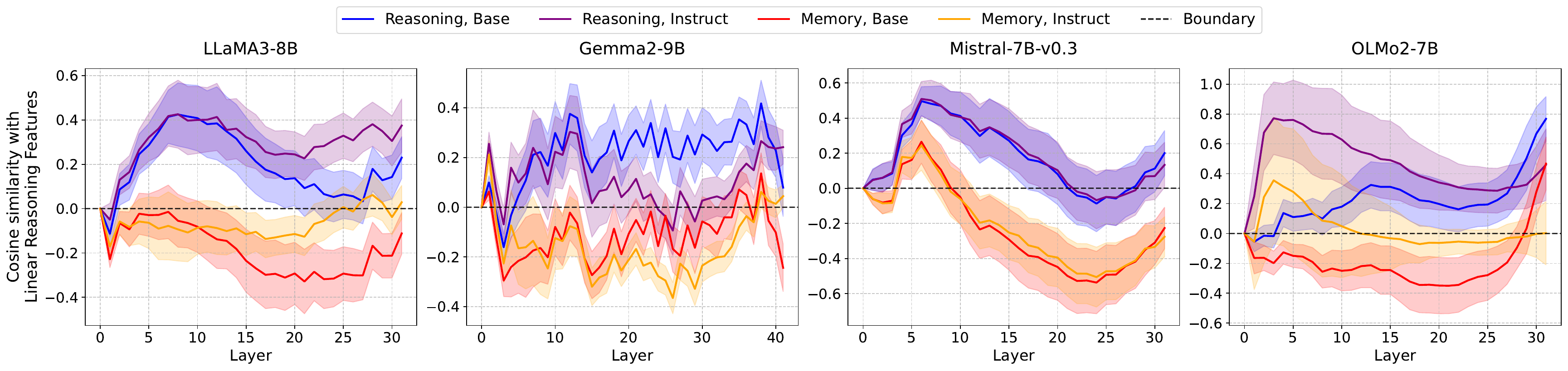}
    \caption{Layerwise cosine similarity between the last token residual stream activations and the extracted Linear Reasoning Features (LiReFs) in four base models and their corresponding instruction-tuned variants.}
    \label{fig: cosine_on_reason_features_with_activations}
\end{figure*}

To quantitatively measure the relation between LiReF and the reasoning capability required for answering each question, we compute the layerwise cosine similarity between the last question token representation of each question and the corresponding LiReF, as shown in Figure \ref{fig: cosine_on_reason_features_with_activations}. For each LLM, we also replicate the same analyses for its pre-trained base version before instruction fine-tuning. 
A positive cosine similarity suggests a positive activation value along LiReF and vice versa. 
We observe that for all eight models, questions in $\mathcal{D}_\textbf{Reasoning}$ mostly activate the reasoning features positively, while questions in $\mathcal{D}_\textbf{Memory}$ mostly have negative LiReF activations, especially in the middle layers. Furthermore, on 3 out of 4 LLM families (LLaMA3-8B, Gemma2-9B, and Mistral-7B-v0.3), the layerwise cosine similarity profiles between the base and instruction-tuned models are highly consistent with each other, suggesting that LLMs may have developed linear reasoning features to mediate its emergent reasoning capability during pre-training rather than post-training. 

% the trend differences between the base models and their instruction-tuned variants are not significant. This further demonstrates that the model's differentiation between memory and reasoning abilities exists even in the base models, stemming inherently from pretraining rather than the post-training process.
% Specifically, in the earlier middle layers, the cosine similarity for reasoning peaks at positive extremes (e.g., the 7th to 10th layers for LLaMA and Mistral, reaching 0.4 and 0.5), while in the later middle layers, Memory’s cosine similarity peaks at negative extremes (e.g., the 20th to 25th layers for LLaMA and Mistral, reaching -0.3 and -0.5), highlighting the most distinct features. 

\subsection{The gradient nature of reasoning-memorization interplay}

%在这里我们需要同时把区域特征分析和error case分析给加进去

\begin{figure}[t]
    \centering
    \includegraphics[scale=0.22]{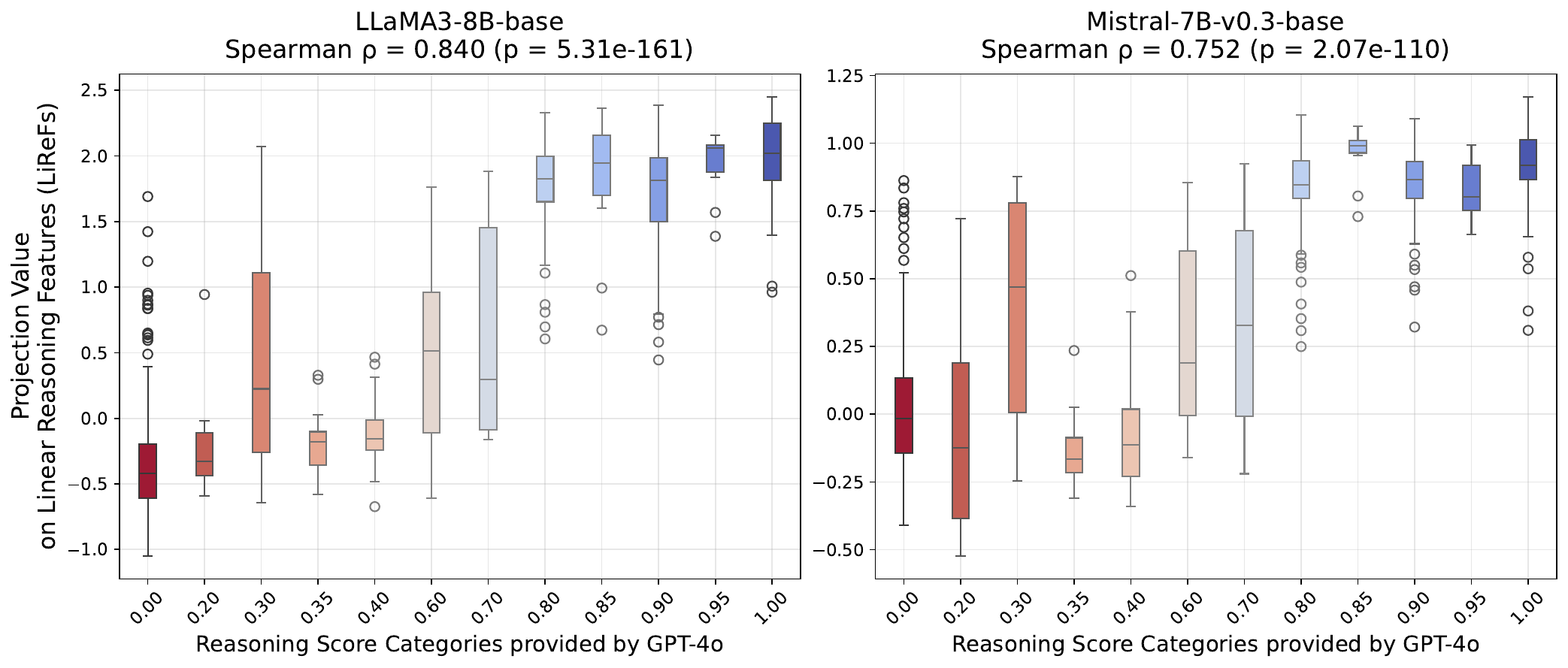}
    \caption{Strong correlation between Projection Values on the Linear Reasoning Features (LiReFs) direction and the Reasoning Score provided by GPT-4o, with Spearman coefficients of 0.840 (LLaMA3-8B-base) and 0.752 (Mistral-7B-v0.3-base). The LiReFs projections exhibit a spectrum-like distribution, where continuous increases in Reasoning Scores correspond to progressively rising Projection Values along the LiReFs direction.}
    \label{fig:specturm}
\end{figure}

% 介于我们在Figure 3中观察到了 belong to $\mathcal{D}_\textbf{Memory}$ 和 $\mathcal{D}_\textbf{Reasoning}$ 的datapoints自然 归属到了boundary的两极的特征，因此我们自然会很好奇，那么它们中间的这段过渡的区域部分代表了什么类型的datapoints，是否在这该direction方向上处于中间位置会代表目标问题需要同时使用了Memory和reasoning两种能力呢？我们通过Figure \ref{fig:specturm}所展示的实验来研究。

As observed in Figure \ref{fig: pca_on_reason_memory}, questions in 
$\mathcal{D}_\textbf{Memory}$ and $\mathcal{D}_\textbf{Reasoning}$ tend to have significantly negative and positive activations along LiReFs, respectively. This raises the question: what types of questions fall near the reasoning-memorization boundary (i.e., those with near-zero LiReF activation values)?
Do these problems require both memory and reasoning abilities to solve?
We investigate this question through the following experiments.

Figure \ref{fig:specturm} shows the relation between GPT-4o-assigned reasoning scores for each question in MMLU-Pro, as discussed in Section \ref{sec: datasets and models}, versus the LiReF projection value $\hat{\mathbf{r}}^T\mathbf{h}^{(l)}(x)$ of its residual stream representation $\mathbf{h}^{(l)}(x)$ by LLaMA3-8B-base and Mistral-7B-v0.3 models. We observe that as problems receive higher reasoning scores assigned by GPT-4o, they tend to have larger activation values along the LiReF direction. This correlation is notably strong across both models, with Spearman correlation coefficients of 0.840 for LLaMA3-8B-base and 0.752 for Mistral-7B-v0.3-base. These findings suggest that problems with near-zero LiReF activations likely involve both memory and reasoning capabilities. 

\begin{figure}[t]
    \centering
    \includegraphics[scale=0.31]{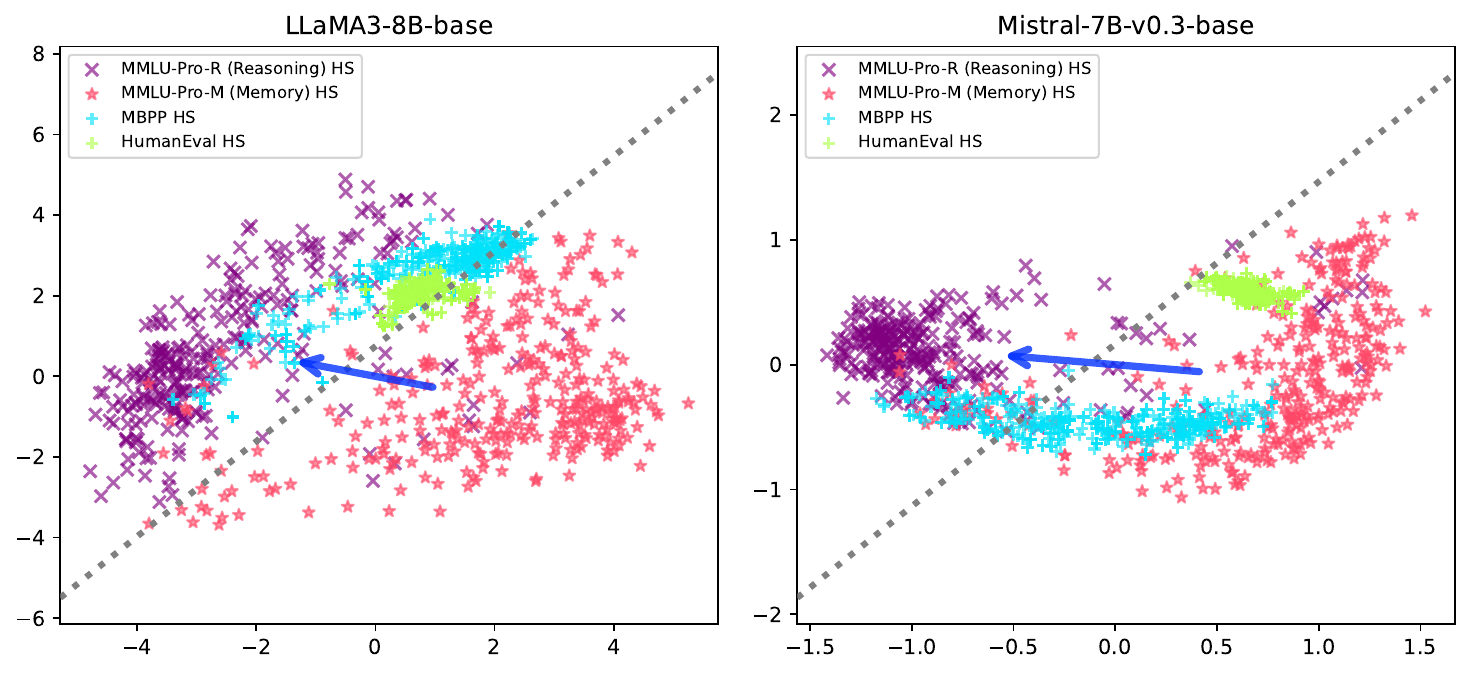}
    \caption{Visualization of the hidden states of two base models on the datasets of MBPP, HumanEval, MMLU-Pro-M and MMLU-Pro-R using 2-dimensional PCA. The hidden states of coding tasks, which involve both reasoning and memory recall, are positioned around the boundary (grey dashed line) fitted via logistic regression.}
    \label{fig:coding_pca}
\end{figure}

To further validate our results, we conducted additional PCA experiments on the Coding tasks - which have been identified by numerous studies as a representative task type requiring both memory and reasoning capabilities in LLMs \citep{zhao2025unveiling, chen2024teaching}. The results are shown in Figure \ref{fig:coding_pca}, where we observe that the residual stream activations of two Coding tasks, MBPP \citep{austin2021programsynthesislargelanguage} and HumanEval \citep{chen2021evaluatinglargelanguagemodels}, are both positioned near the boundary. This further supports our finding that data points situated between the two extremes represent task types that engage both memory and reasoning abilities in LLMs.

% 后面再加一个其他的Coding datasets的points

% 为了更好地印证这一点，我们对Coding任务类型的数据， which have been provded by 好一些文章that 是一个非常代表性的LLM同时使用Memory and Reasoning的任务，进行PCA实验，结果如Figure5所示。

% \paragraph{Error Cases Analysis}

% In \S\ref{sec: causal validation}, we will try to 
% 调节Linear reasoning features，来纠正这些错误的cases，让他们往更适合自己的features区间去靠近

% the spectrum distribution along the LiReFs direction naturally raises the question of whether pushing the model's representation further into the reasoning region would lead to better reasoning performance in certain tasks. We will explore this in Section \ref{sec: causal validation} through Features Intervention experiments.

\section{Causal validation of LiReFs}
\label{sec: causal validation}

% 在这个section，我们通过对模型的hiddenstatesxxx 进行intervention的实验，来因果验证xxx

%MMLU-pro 5 shot
%gms8k 8 shot

\subsection{Inference-time LiReF intervention}
\label{sec: intervention}

\begin{table*}
\setlength{\belowcaptionskip}{-10pt}
\centering
\resizebox{1.0\linewidth}{!}{
\begin{tabular}{l c c c c c c}\\
\toprule
& \multicolumn{3}{c}{Memory-Intensive Datasets} & \multicolumn{3}{c}{Reasoning Datasets} \\
\cmidrule(lr){2-4} \cmidrule(lr){5-7}
Base Model & \text{MMLU-Pro-M} & \text{PopQA} & \text{C-Eval-H} & \text{MMLU-Pro-R} & \text{GSM-8k} & \text{MGSM} \\
\midrule
\text{LLaMA3-8B-base} & \gray{41.1} / \textbf{48.3} \ua{7.2} & \gray{33.4} / \textbf{35.6} \ua{2.2}  & \gray{45.2} / \textbf{47.4}  \ua{2.2} & \gray{24.2} / \textbf{33.5} \ua{9.3}  & \gray{49.0} / \textbf{53.1} \ua{4.1} & \gray{28.5} / \textbf{34.6} \ua{6.1}   \\
\text{Gemma2-9B-base} & \gray{37.5} / \textbf{50.1} \ua{12.6} & \gray{29.2} / \textbf{30.3} \ua{1.1}  & \gray{52.1} / 52.1 & \gray{29.2} / \textbf{44.7} \ua{15.5} & \gray{61.9} / \textbf{63.5} \ua{1.6} & \gray{45.8} / \textbf{47.0} \ua{1.2}  \\
\text{Mistral-7B-v0.3-base} & \gray{37.8} / \textbf{43.6} \ua{5.8} & \gray{30.1} / \textbf{30.9} \ua{0.8} & \gray{38.2} / \textbf{44.0} \ua{5.8} & \gray{20.8} / \textbf{21.7} \ua{0.9} & \gray{35.1} / \textbf{36.2} \ua{1.1} & \gray{12.0} / 12.0 \\
\text{OLMo2-7B-base}  & \gray{19.4} /  \textbf{25.0} \ua{5.6} & \gray{19.2} / \textbf{20.1} \ua{0.9} & \gray{26.0} / \textbf{28.9} \ua{2.9} & \gray{11.3} / \textbf{16.5} \ua{5.2} & \gray{11.5} / \textbf{12.3} \ua{0.8} & \gray{10.1} / \textbf{11.3} \ua{1.2} \\
\bottomrule
\end{tabular}}
\caption{
The performance of four base models on six benchmarks, before and after feature intervention. The results indicate that by shifting the residual stream of the reasoning-required or memory-required tasks further to the specific feature regions, overall task performance can be substantially enhanced.}
\label{tab: intervention results}
\end{table*}

% In this section, 我们会尝试 人为地在 Inference-time的阶段对模型内部的residual stream activations 进行 Intervene的实验, 来观察其对 模型在Memory-Intensive tasks和 Reasoning Tasks上各自的表现影响。
In this section, we conduct experiments where we manually intervene in the residual stream activations during inference time. By adjusting the intensity of linear reasoning features in model residual streams, we examine how model performance on both memory-intensive and reasoning-intensive tasks will change.

In particular, for all tokens of each question, we modify their residual stream representations in a specific layer by adding an intervention vector along the LiReF direction, as suggested in Equation \ref{eq:feature-addition}. To enhance the most relevant model capability, we adopt negative values of $\alpha$ for $\mathcal{D}_\textbf{Memory}$, and positive $\alpha$ values for $\mathcal{D}_\textbf{Reaoning}$.  After carefully tuning $\alpha$ on validation sets, we ask each model to generate answers for questions in $\mathcal{D}_\textbf{Memory}$ and $\mathcal{D}_\textbf{Reaoning}$, and measure its performance change under inference-time LiReF intervention. More details about the experimental setup, including the validation-test set splits, hyperparameter selection criteria and inference settings can be found in Appendix~\ref{appendix: details of intervention experiments}.
% The experiments are conducted across the same six datasets and four base models as described in Section \ref{sec: datasets and models}.
% More details about the experimental setup, including the validation-test set splits, hyperparameter selection criteria and inference settings can be found in Appendix~\ref{appendix: details of intervention experiments}.

The main results are shown in Table \ref{tab: intervention results}. We observe that intervening LiReFs during inference time effectively improves the performance of four LLMs on both memory-intensive and reasoning-intensive tasks. Moreover, the improvements remain consistent across different task types, domains, and languages, further supporting our claim that the reasoning features in  LLM residual streams capture general reasoning capability. In the next section, we will present specific cases to illustrate how reasoning feature intervention improves model performance by reducing reasoning step errors and correcting the misapplication of model abilities.

% We show that LiReFs can be leveraged to reduce LLM reasoning errors and detect data contamination caused by model memorization.

%gemma好像不太适合做多选题，更适合 open-ended generation？

% & 51.8 {\color{baseline_color} / 51.7} \colornumber{(+0.1)} & 51.7 {\color{baseline_color} / 51.5} \colornumber{(+0.2)} & 31.3 {\color{baseline_color} / 32.0} \colornumber{(-0.7)} & 44.7 {\color{baseline_color} / 47.1} \colornumber{(-2.4)}

% \begin{equation}
% \label{eq:proj}
%     \mathbf{v}_{\perp} = \mathbf{v} - \text{proj}_{\mathbf{d}}(\mathbf{v}) = \mathbf{v} - \frac{\mathbf{v} \cdot \mathbf{d}}{\|\mathbf{d}\|^2} \mathbf{d},
% \end{equation}

\subsection{Cases Study}
\label{sec: cases analysis}

In the PCA analyses presented in Section \ref{sec: robustness analysis}, we observed certain sample cases that, although labeled as reasoning-intensive by GPT-4o or by the task name, have negative LiReF activations on the memorization subspace. Similarly, some cases that were labeled as memory-intensive instead fall into the reasoning subspace with positive-valued LiReFs. In this section, we analyze these cases and also conduct LiReF intervention experiments, aiming to correct any potential reasoning errors or unfaithful reasoning steps.
% In this section, we analyze these specific cases 

% and then also conduct feature intervention experiments to interfere with their inference process, aiming to correct any potential reasoning errors or unfaithful reasoning steps.
% Firstly, we use LiReFs and the Reasoning score to identify these cases, and then perform a manual selection to obtain a small dataset, approximately 100 to 200 samples. 
% After that, we conduct the intervention and compare the accuracy before and after the intervention.

Firstly, we collect questions in MMLU-Pro whose reasoning label contradicts the actual feature subspace in which they are positioned. (e.g., cases whose GPT-4o-assigned reasoning score is much less than 0.5, but have a positive-valued LiReF activation), and evaluate LLaMA3-8B-base on them to identify a subset of questions where the model provides incorrect answers. 
Then we obtained a subset of 184 cases in total. 
Next, we perform inference-time LiReF intervention on these examples following the same settings in Section \ref{sec: intervention}, and compare their accuracy and actual outputs before and after the intervention.
We found that, by shifting LiReF activation to have the sign that is consistent with GPT-4o-assigned reasoning score, model accuracy on this subset jumps from 0 to 0.21. Table \ref{tab:cases examples} presents some exemplar questions in our analyses, together with model answers before and after LiReF intervention. These results suggest that LLM reasoning errors might not be due to a lack of relevant knowledge, but are caused by the insufficient activation of its acquired generalizable thinking capabilities, which can be alleviated through targeted inference-time intervention of reasoning features.
% 并且恢复了正确答案.
% 我们可以看到，当对一些

% Taken together, our analyses suggest that model 

% model’s poor performance is not necessarily due to a lack of relevant knowledge or reasoning ability, but rather because it may have adopted the wrong approach to solving the problem. By correcting the reasoning process through Reasoning Features intervention, the model is able to arrive at the correct answer on its own, without the need for additional external knowledge.

% 经过Interven之后，他们从Memor-Recall的思维方式回复到了 Reasoning主导的思维方式，并且最终给出了正确答案

% 需要提到 overthinking的问题

\begin{table*}[t]
    \centering
    \resizebox{\textwidth}{!}{
    \scriptsize
    \setlength{\tabcolsep}{4pt}
    \begin{tabular}{p{3.2cm}|p{4.2cm}|p{4.2cm}|p{1.4cm}}
    \toprule
        \bf Example Question & \bf Answer before Intervention & \bf Correct Answer after Intervention & \bf Error Reason \\
    \midrule
        Speed of data transmission in 4-G network of telecom is? & 4G networks are designed to provide ultra-fast speeds, and with advancements in LTE-Advanced and carrier aggregation, they can reach speeds as high as 10 Gbps. The answer is \textbf{10 Gbps}.	 & The speed of data transmission in a 4G network typically ranges from 100 Mbps (megabits per second) to 1 Gbps (gigabit per second), depending on factors such as network congestion. The answer is \textbf{100 Mbps to 1 Gbps}.	 & Shallow Memory recall \\
    \midrule
        A sound wave with frequency f travels through air at speed v. With what speed will a sound wave with frequency 4f travel through the air? & 4f is four times the frequency of f, so the wavelength of 4f is one-fourth the wavelength of f. Since the speed of sound is the product of frequency and wavelength, the speed of 4f is one-fourth the speed of f. The answer is \textbf{v/4}. & The speed of sound in air is independent of the frequency of the sound wave. It only depends on factors like the medium and the temperature of the air. So, if a sound wave with frequency f travels through air at speed v, a sound wave with frequency 4f will also travel through the air at the same speed, v. The answer is \textbf{v}. & Overthinking  \\
    % \midrule
    %     In Harry Potter, what is the name of Hagrid's giant spider friend? & Hagrid's giant spider friend is named \textbf{Aragog}. &  \textbf{Aragog} \\
    \bottomrule
    \end{tabular}
    }
    \caption{Example questions and outputs before and after features intervention on the LLaMA3-8B-base model. They are both Memory-Intensive questions, but the features were incorrectly positioned in the Reasoning region. After shifting the Features to the appropriate region, the error was corrected.}
    \label{tab:cases examples}
\end{table*}

% 都是因为是Mmeory的问题但Features却在Reasoning的Region当中导致了错误，通过shift features到合理的区域后，错误得到纠正

% 这里需要有一个table

% 首先还是需要借助LiReFs 和 Reasoning score来找出这些cases，然后来进行认为筛选，得到一小批数据，大概100～200个？然后做interven，比较前后准确率

% We show that linear reasoning feature can be leveraged to reduce LLM reasoning errors, detect unfaithful reasoning steps, and mitigate reasoning data contamination caused by memorization (Section 5).

% 那些错误地走向memory or reasoning的cases，将其推向反方向reasoning or memory，会不会拥有更高准确率呢 （缓解over-thinking or 其他问题）
% 首先还是需要借助LiReFs 和 Reasoning score来找出这些cases，然后来进行认为筛选，得到一小批数据，大概100～200个？然后做interven，比较前后准确率

\subsection{Reasoning Generalization Effects}
\label{sec: reasoning generalization}

%开始还是可以写PCA的实验以及special cases的实验发现可能是这样的例子。然后进一步scale up，研究在这个direction上的intensity对模型reason泛化性的表现影响效果。

% Here, in order to better evaluate the performance of xxxx, we design a reasoning dataset that xxxxx
% , in order to better evaluate the generlization of model reasoning ablity, avoiding the cheating from memory...

\begin{figure*}[t]
\setlength{\belowcaptionskip}{-8px}
    \centering
    \includegraphics[scale=0.32]{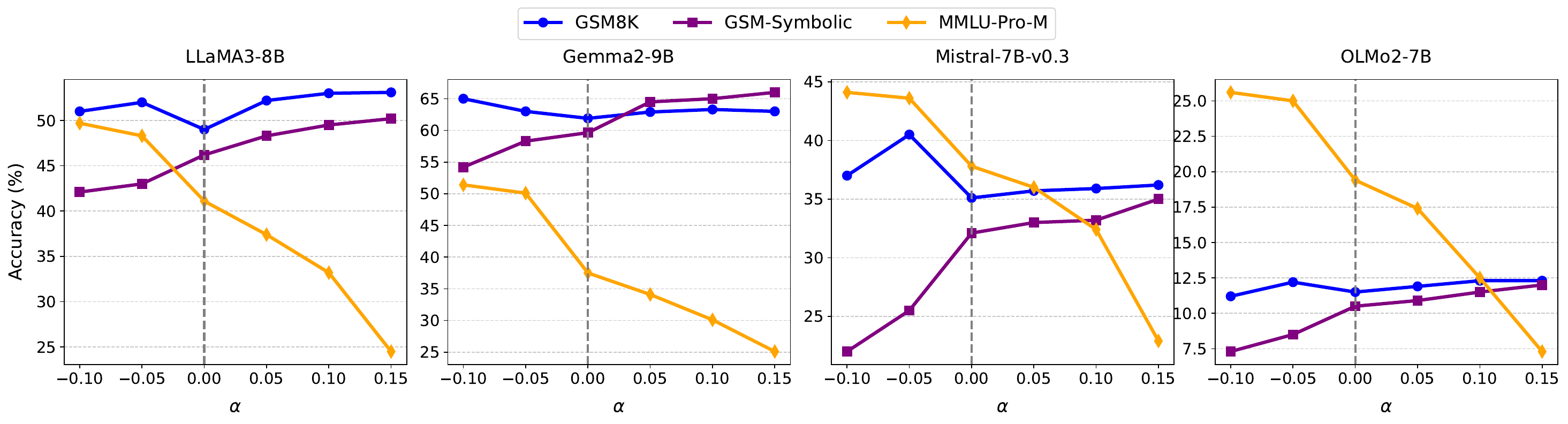}
    \caption{Performance of the four base models on the GSM-8k, GSM-Symbolic, and MMLU-Pro-M datasets, with varying hyperparameter $\alpha$ to control the intensity of feature intervention, shows that all four models exhibit potential data leakage risks on the GSM-8k dataset. The models may rely on memory to achieve good performance on this reasoning task.}
    \label{fig: generalization_performance}
\end{figure*}

In the previous experiments, we noticed that the features of certain questions from reasoning datasets lie in the memory subspace with negative LiReF activations. Therefore, we suspect that the models might have solved these reasoning questions through memorization (possibly due to training data contamination), rather than applying genuine reasoning capability that is generalizable under systematic input variation. To verify this hypothesis, we conduct additional features intervention experiments on GSM-Symbolic \citep{mirzadeh2025gsmsymbolic} in this section. GSM-Symbolic is a variant of GSM-8k. It selects 100 question templates from GSM-8k and then generates 50 different instances for each template by varying numerical conditions, results, and other factors. The resulting dataset contains 5,000 data points, making it ideal for a reliable evaluation of the model's reasoning generalization capabilities.

% 这里非常非常重要，要讲清楚，泛化性表现越好的samples，是否会在LiReFs方向上有更大的投影值
% 我们怀疑这样的cases是，而缺乏泛化性，而事实也正如我们所示

% 假若问题中的某个条件 or 数值发生变化
% 因此我们在这部分中，我们选择了
Figure \ref{fig: generalization_performance} shows mean model accuracy on GSM-Symbolic, GSM-8k, and MMLU-Pro-M under inference-time LiReF intervention. We can see that as the intervention intensity $\alpha$ increases from 0, the performance of all four models on both GSM-8k and GSM-Symbolic rises consistently. On the other hand, as $\alpha$ decreases from 0, we observe that, compared to GSM-8k, GSM-Symbolic experiences a more significant performance drop with suppressed LiReFs. Notably, the performance gain and loss on GSM-Symbolic suggests that LiReF intervention is likely enhancing the genuine model reasoning capability that is generalizable, as opposed to case-based reasoning skills that rely more on memorization of particular training examples. Interestingly, we also observe that the performance drop on GSM-8K under LiReF suppression is less pronounced compared to GSM-Symbolic, and there is even a slight improvement with a moderate suppression when setting $\alpha = -0.05$. This implies that the model might have previously been exposed to GSM-8K due to data leakage, and therefore adopts a memory-intensive strategy to answer these questions. While for MMLU-Pro-M, performance improves steadily as $\alpha$ decreases, supporting our observation that reducing the projection values of the model’s residual stream in the LiReFs direction enhances the model's ability to recall from memory.

% This confirms our previous finding that shifting the residual stream to increase the their projection values on LiReFs direction will enhance the models' reasoning capabilities. Moreover, the improvement on GSM-Symbolic suggests that 

% this enhancement is not due to the model's memory, but rather a true increase in the model's reasoning generalization ability (since the model cannot have memory of the data from GSM-Symbolic). As $\alpha$ decreases from 0, we observe that, compared to GSM-8k, GSM-Symbolic experiences a more significant drop in performance, indicating that the model's performance on GSM-Symbolic relies more on genuine reasoning ability. In contrast, the drop in performance on GSM-8k is less pronounced, and there is even a slight improvement when $\alpha = -0.1$. This implies that the model might be benefiting from data leakage in GSM-8k, which allow it to rely on existing memory to answer questions and maintain performance. On the other hand, for MMLU-Pro-M, performance improve steadily as $\alpha$ decreases, supporting our observation that reducing the projection values of the model’s residual stream in the LiReFs direction enhances the model's ability to recall from memory.

% 此外， 我们怀疑在Reasoning-Memorization Sepcturm中， 越接近reasoning features region越深的地方，往往能够调动模型更强的泛化性。 然后我们做了观察本身的那个数据点的表现实验，证实了这个

% 存留的GSM-8k记忆

% \subsection{Reasoning circuit fine-tuning}

% \input{paper/5_application}
\section{Conclusion}
% In this study, we show that the reasoning and memorization abilities of LLMs across various domains and languages are driven by a shared set of linear reasoning features within their activation space. These features can be uncovered by contrasting the hidden representations of reasoning-intensive and memory-intensive queries. Furthermore, we demonstrate that intervening on these Linear Reasoning Features (LiReFs) during model inference improves both accuracy and robustness to perturbations when handling reasoning tasks. We also reveal that manipulating these reasoning features enables the model to more precisely activate the most relevant problem-solving capabilities during answer generation. Our findings offer a mechanistic understanding of how reasoning and memorization interact in LLMs, providing a foundation for future research aimed at building more capable and interpretable generative reasoning systems. 

In this study, we show that the reasoning and memorization abilities of LLMs across various domains and languages are driven by a shared set of linear reasoning features within their activation space. These features can be uncovered by contrasting the hidden representations of reasoning-intensive and memory-intensive queries. Furthermore, we demonstrate that intervening on these Linear Reasoning Features (LiReFs) during model inference improves both accuracy and robustness to perturbations when handling reasoning tasks. We also reveal that manipulating these reasoning features enables the model to more precisely activate the most relevant problem-solving capabilities during answer generation. Our findings offer a mechanistic understanding of how reasoning and memorization interact in LLMs, providing a foundation for future research aimed at building more capable and interpretable generative reasoning systems. Additionally, our work suggests that enhancing the interpretability of these underlying features could lead to more efficient and focused interventions, contributing to the development of models that are both more powerful and more transparent in their decision-making processes.
\section*{Limitations}
Our work has several limitations. First, we only studied reasoning features in relatively small LLMs, while recent studies show that by scaling up both model size and inference-time computation, the reasoning capability of LLMs can be significantly improved \citep{hoffmann2022training,openai2024learning}. Second, we have focused mostly on reasoning problems that can be addressed through short answers, while it remains unclear whether LiReFs can be utilized to enhance model's ability of performing deliberate reasoning via various prompt engineering techniques such as chain-of-thought \citep{wei2022chain}, self-reflection \citep{shinn2024reflexion}, and tree-of-thought \citep{yao2024tree}. Third, we formulate memorization as performance inconsistency against reasoning question perturbation, while another line of LLM reasoning research has employed a different definition of \textit{counterfactual memorization} -- i.e., change of model answers on particular test questions after removing a similar example from training data \citep{zhang2023counterfactual,hu2024case}. Future work should investigate Whether perturbational and counterfactual memorization are mechanistically equivalent and, therefore, can be both mediated by LiReFs. 

\section*{Acknowledgment}
This material is based in part upon work supported by Schmidt Sciences; by the German Federal Ministry of Education and Research (BMBF): Tübingen AI Center, FKZ: 01IS18039B; by the Machine Learning Cluster of Excellence, EXC number 2064/1 – Project number 390727645. 


\bibliography{acl_latex}

\clearpage
\appendix

\section{Prompts}
\label{appendix: prompt}

Table \ref{tab: prompt} presents the prompt we used to query GPT-4o to assign a Reasoning Score to each question.

\begin{table*}[htbp]
\centering
\begin{examplebox}
\ttfamily
\begin{itemize}
    \item \textbf{Analyze the question to determine its position on the reasoning-memory spectrum. Return:}
    \begin{enumerate}
        \item Concise justification (1-2 sentences)
        \item Score [0.0--1.0] where:
        \begin{itemize}
            \item 1.0 = Strictly requires multi-step reasoning (calculations/formulas/deductions)
            \item 0.0 = Purely factual recall or the inference of humanities knowledge
            \item Intermediate values indicate hybrid characteristics
        \end{itemize}
    \end{enumerate}

    \textbf{Scoring Guidelines:}
    \begin{itemize}
        \item +0.5 if contains numerical values/percentages
        \item +0.3 per required calculation step
        \item +0.2 if requires unit conversions
        \item -0.4 if answer appears verbatim in STEM textbooks
        \item Max 1.0 | Min 0.0
    \end{itemize}

    \textbf{Examples:}
    \begin{enumerate}
        \item \textbf{Score 0.0:}
        \begin{quote}
            Question: ``Polarization is a property of...'' \\
            Options: [transverse waves,...] \\
            Analysis: Directly tests textbook knowledge about wave properties without calculations. \\
            \textbf{Score: 0.0}
        \end{quote}

        % \item \textbf{Score 0.95:}
        % \begin{quote}
        %     Question: ``Clay contains 30\% AI₂O₃... limestone required?'' \\
        %     Options: [4.80×10⁶ grams...] \\
        %     Analysis: Requires multi-step stoichiometric calculations (mass\% → moles → equation ratios → unit conversions). \\
        %     \textbf{Score: 0.95}
        % \end{quote}

        \item \textbf{Score 0.35:}
        \begin{quote}
            Question: ``An owner of an apartment building in a rundown section of town knew...If the neighbor asserts a claim against the owner to recover damages for his injury, he should'' \\
            Options: [not recover, because the owner can't be held responsible...] \\
            Analysis: Humanities-oriented question, which, although requiring multi-step reasoning, still leans more towards a memorization-based approach. \\
            \textbf{Score: 0.35}
        \end{quote}

        % \item \textbf{Score 0.85:}
        % \begin{quote}
        %     Question: ``Energy of photon with 500nm wavelength?'' \\
        %     Options: [3.97eV,...] \\
        %     Analysis: Requires formula (E=hc/λ) and single-step calculation with standard constants. \\
        %     \textbf{Score: 0.85}
        % \end{quote}

        \item \textbf{Score 0.95:}
        \begin{quote}
            Question: ``Order from greatest to least: 3, 3 and 1 over 8, 3.8, 3.18.'' \\
            Options: ['3.8, 3 and 1 over 8, 3.18, 3',...] \\
            Analysis: Requires comparing numerical values and determining their order. \\
            \textbf{Score: 0.95}
        \end{quote}
    \end{enumerate}

    \textbf{Current Analysis:}
    \begin{quote}
        Question: ``\{question\_text\}'' \\
        Options: \{options\_list\} \\
        Analysis: 
    \end{quote}
\end{itemize}

\rmfamily
\end{examplebox}
\caption{Prompt used to query GPT-4o to assign a Reasoning Score to each question.}
\label{tab: prompt}
\end{table*}

\section{Details of Datasets}
\label{appendix: details of datasets}

Here, we provide further details about the datasets used in Sections \ref{sec:liref} and \ref{sec: causal validation}.

\paragraph{MMLU-Pro-M \citep{wang2024mmlupro} and MMLU-Pro-R} MMLU-Pro is a comprehensive benchmark designed to assess the advanced language understanding and reasoning capabilities of large language models (LLMs). It spans 14 diverse domains such as mathematics, physics, chemistry, law, engineering, psychology, and health, encompassing over 12,000 questions. It features 10 options per question, significantly increasing the difficulty and robustness of the benchmark. Unlike MMLU, MMLU-Pro focuses on more challenging college-level problems that require deliberate reasoning across various domains. In this work, we use GPT-4o to assign a Reasoning Score to each question. We then divide the questions into two subsets: those with a score greater than 0.5 are categorized as MMLU-Pro-R, while those with a score of 0.5 or below are classified as MMLU-Pro-M.

\paragraph{PopQA \citep{mallen-etal-2023-trust}} PopQA focuses on evaluating factual knowledge in large language models, specifically targeting knowledge about entities, defined as triplets of (subject, relationship, object). The task is framed as open-domain question answering, where a model is asked to predict an answer without pre-given ground-truth paragraphs. This study explores few-shot learning and prompts LMs without parameter updates, in contrast to fine-tuning approaches. The performance is measured by accuracy, where a prediction is considered correct if any substring matches a gold answer.

\paragraph{C-Eval-H \citep{huang2023ceval}} C-EVAL is a comprehensive Chinese evaluation suite designed to assess the advanced knowledge and reasoning abilities of large language models (LLMs) in a Chinese context. As traditional NLP benchmarks primarily focus on English and fail to capture the unique challenges of Chinese language models, C-EVAL addresses this gap by providing a detailed evaluation framework tailored to the Chinese language and culture. It includes 13,948 multiple-choice questions across 52 diverse disciplines, ranging from humanities to science and engineering, and spans four difficulty levels: middle school, high school, college, and professional exams. In this work, we focus on the humanities portion and refer to it as C-Eval-H.

\paragraph{GSM8k \citep{cobbe2021trainingverifierssolvemath}} GSM8k is a dataset designed to evaluate the mathematical reasoning abilities of large language models (LLMs). It consists of 8.5K grade school-level math problems paired with natural language solutions. The dataset aims to address the challenges faced by LLMs in performing multi-step mathematical reasoning, which often reveals a critical weakness in these models.

\paragraph{MGSM \citep{shi2022languagemodelsmultilingualchainofthought}} The MGSM (Multilingual Grade School Math) benchmark is introduced to assess multilingual reasoning abilities in large language models, addressing the gap between English-based chain-of-thought (COT) reasoning and multilingual NLP tasks. Building on the GSM8K dataset, MGSM extends it to ten typologically diverse languages through manual translations. 
.

% \begin{table*}
% \setlength{\belowcaptionskip}{-10pt}
% \centering
% \resizebox{1.0\linewidth}{!}{
% \begin{tabular}{l l l l l l}\\
% & Dataset & Task & Language & Domain & Examples \\
% \midrule
% \multirow{3}{.1em}{\rotatebox{90}{\textbf{}}} & 
% \text{MMLU-Pro-M} & Multi-Choice Questions Answering & English & History, Law, Politics, etc... & \\
% & \text{PopQA} & Open-ended Generation & English & Factual Knowledge & \\
% & \text{C-Eval-H} & Multi-Choice Questions Answering & Chinese & Humanities & \\
% \midrule
% \multirow{3}{.1em}{\rotatebox{90}{\textbf{}}}
% & \text{MMLU-Pro-R} & Multi-Choice Questions Answering & English & Physics, Chemistry, Biology, etc... & \\
% & \text{GSM8K} & Open-ended Generation & English & Mathematics & \\
% & \text{MGSM} & Open-ended Generation & French, Russian, etc... & Mathematics & \\

% \bottomrule
% \end{tabular}}
% \caption{The list of datasets used in this work.}
% \label{tab:list_of_datasets}
% \end{table*}

\section{Additional Experiments}
\label{appendix: additional experiments}

% \subsection{Verification of the token distributions influence}
% \label{appendix:tok_distribution_verified}

\subsection{Detailed Plot of the PCA results}

\begin{figure*}[t]
\setlength{\belowcaptionskip}{-8px}
    \centering
    \includegraphics[angle=270,scale=0.3]{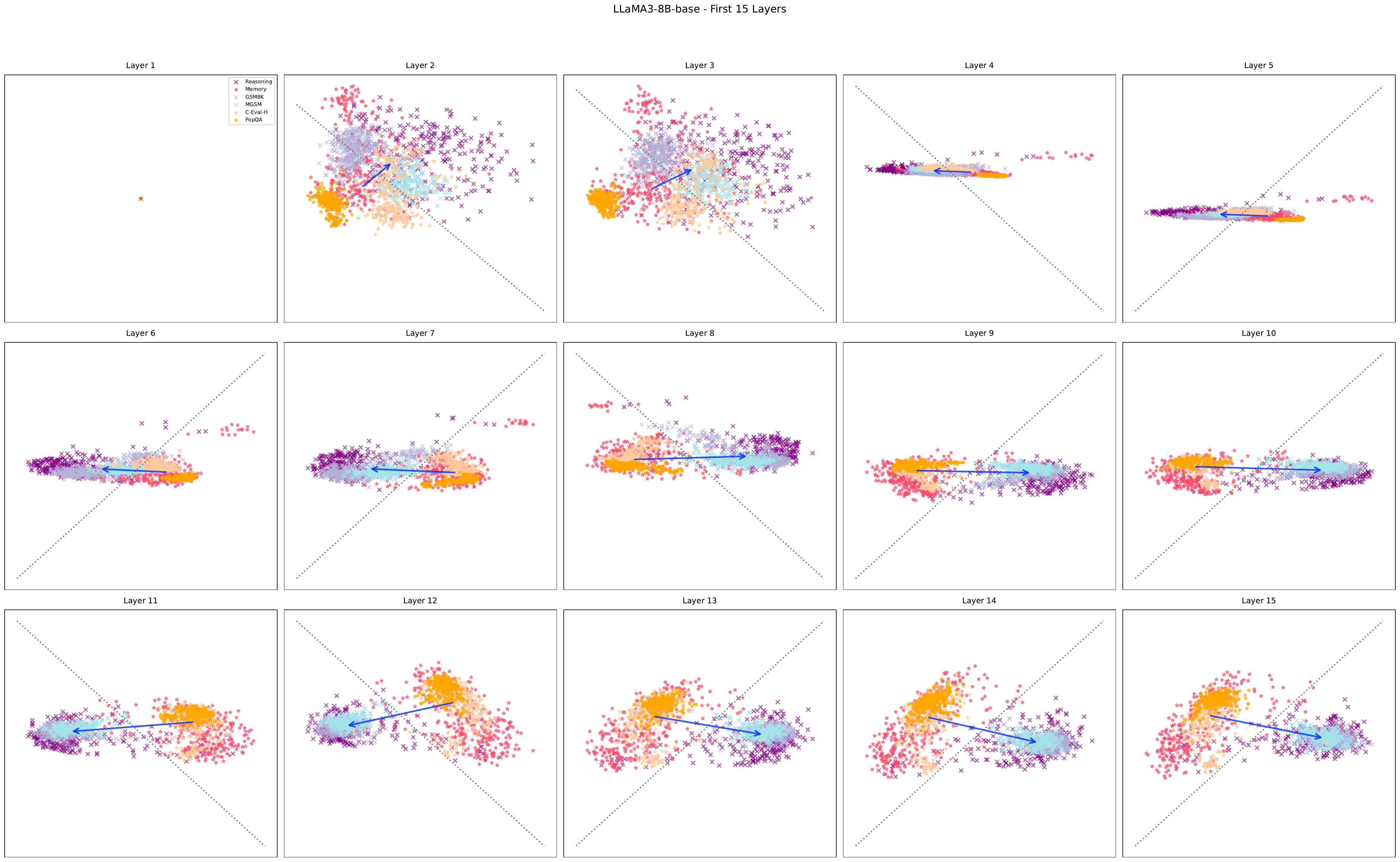}
    \caption{The PCA experiments results on the first 15 layers on LLaMA3-8B-base models}
    \label{fig: more pca layers on llama}
\end{figure*}

\begin{figure*}[t]
\setlength{\belowcaptionskip}{-8px}
    \centering
    \includegraphics[angle=270,scale=0.3]{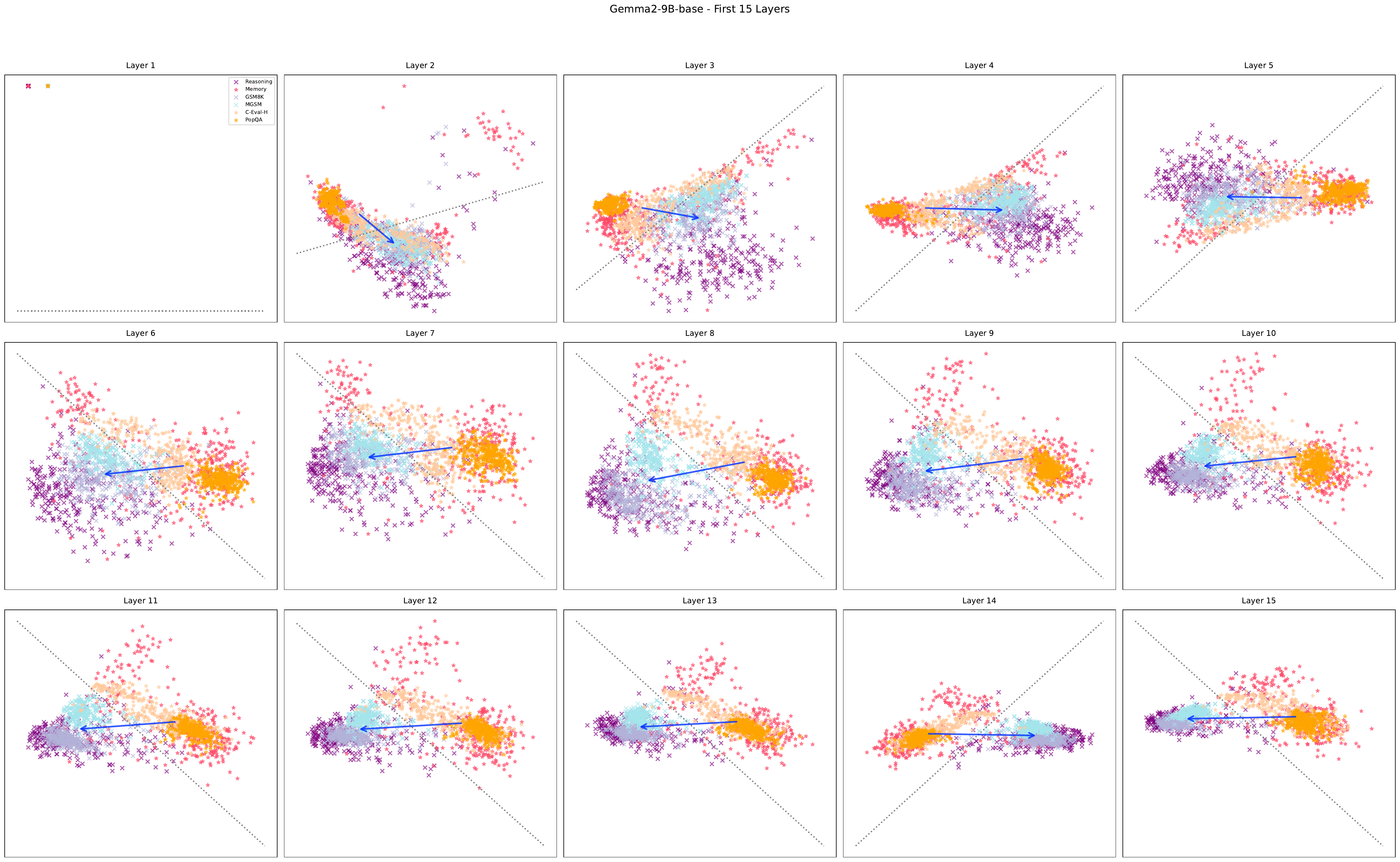}
    \caption{The PCA experiments results on the first 15 layers on Gemma2-9B-base models}
    \label{fig: more pca layers on gemma}
\end{figure*}

\begin{figure*}[t]
\setlength{\belowcaptionskip}{-8px}
    \centering
    \includegraphics[scale=0.29]{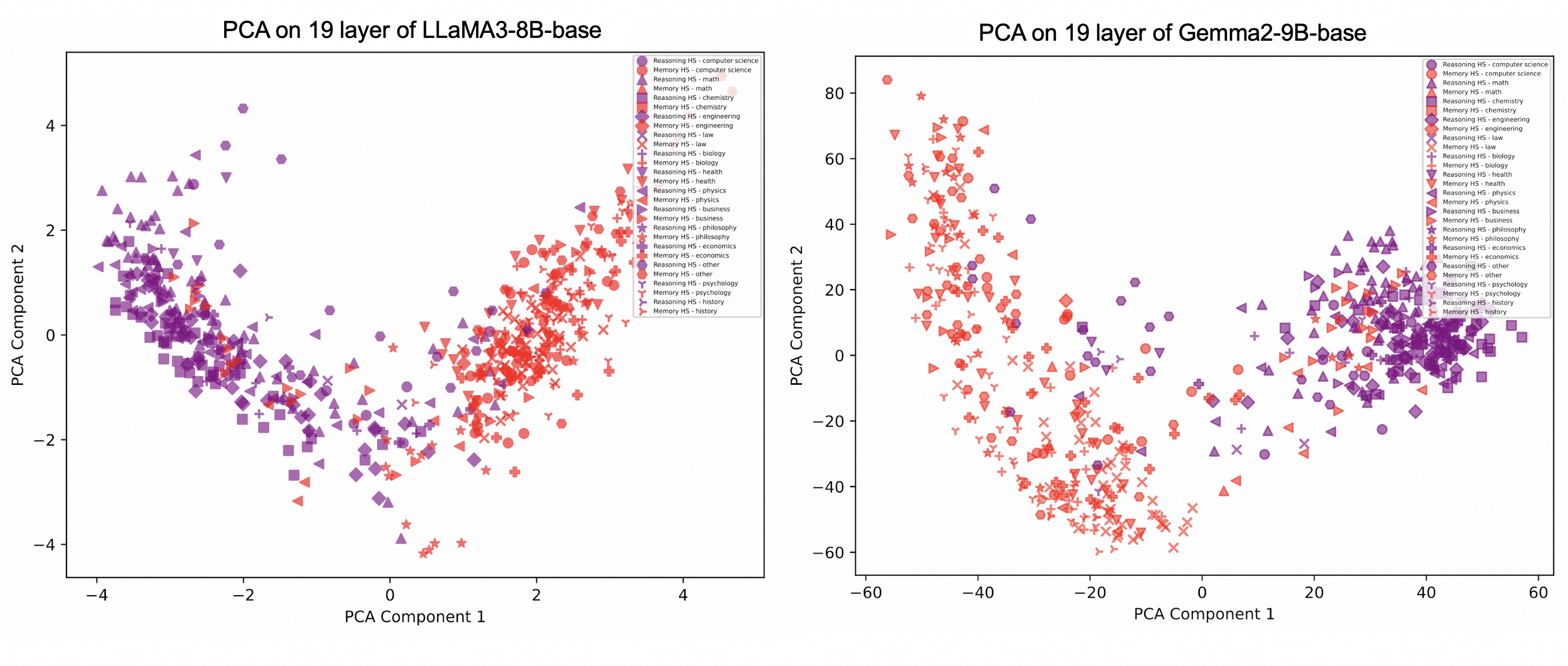}
    \caption{Fine-grained PCA visualizations of questions from different subject domains in MMLU-Pro on the model of LLaMA3-8B-base and Gemma2-9B-base.}
    \label{fig: fined grained}
\end{figure*}

\begin{figure*}[t]
\setlength{\belowcaptionskip}{-8px}
    \centering
    \includegraphics[scale=0.29]{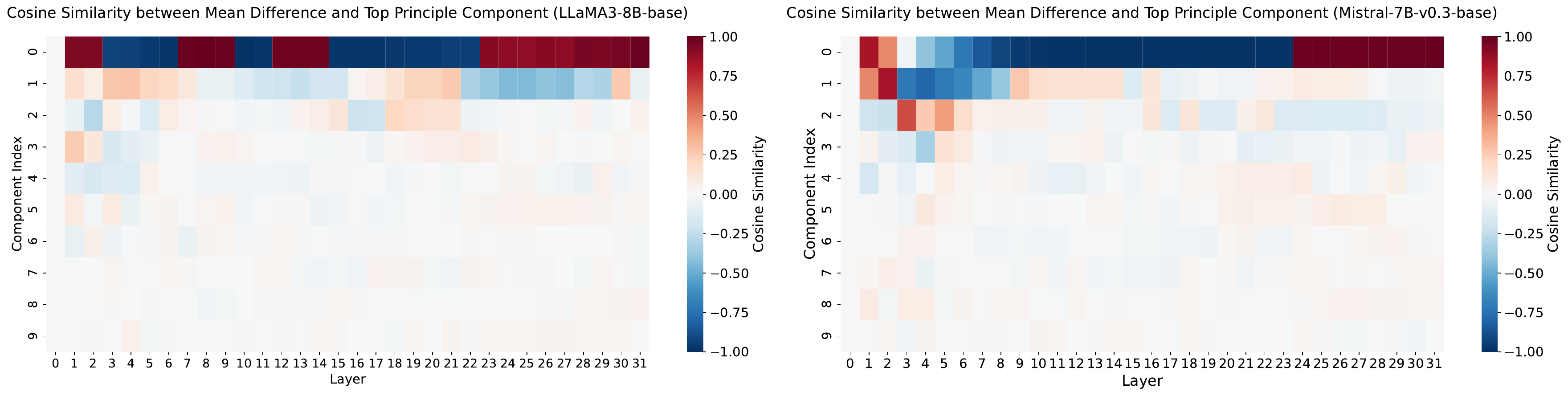}
    \caption{The top one principal component in PCA experiments captures most of the mean difference (Equation \ref{eq:hh-refusal-feature} between the activations in $\mathcal{D}_\textbf{Memory}$ and $\mathcal{D}_\textbf{Reasoning}$.}
    \label{fig: cosine_pca_mean_diff}
\end{figure*}

In this section, we present additional PCA results from various layers of the LLaMA3-8B-base and Gemma2-9B-base models discussed in Section \ref{sec: datasets and models}, which is shown in Figure \ref{fig: more pca layers on llama} and Figure \ref{fig: more pca layers on gemma}. We also provide fine-grained PCA visualizations of questions from different subject domains in MMLU-Pro in Figure \ref{fig: fined grained}. Additionally, we include heatmaps in Figure \ref{fig: cosine_pca_mean_diff} demonstrating that the first principal component from our PCA experiments captures the majority of the mean activation differences between $\mathcal{D}_\textbf{Memory}$ and $\mathcal{D}_\textbf{Reasoning}$.

% \section{Hyperparameters Analysis}
\section{Details of the Intervention Experiments}
\label{appendix: details of intervention experiments}

% We follow the original experimental setup for all the datasets used, e.g the GSM8K and MGSM are run in 8 shots

Here, we provide more implementation details in the Features Intervention Experiments described in Section \ref{sec: causal validation}. 

\paragraph{Inference Settings}

For the few-shot settings, we adhere to the original experimental setup across all datasets. Specifically, we use 5-shot for MMLU-Pro-M, MMLU-Pro-R, and C-Eval-H, and 8-shot for GSM8k, MGSM, and GSM-Symbolic. Additionally, we run 0-shot for PopQA, following the original configuration.

For both open-ended generation and multi-choices question answering tasks, we allow the model to generate the next 200 tokens.

\paragraph{Validation-Test Set Split}

% 我们直接使用各个数据集本身所切分好的validation set 和 test set来进行inference
For parameter tuning and inference, we directly utilized the pre-existing validation and test sets that were already partitioned within each dataset.

\paragraph{Hyperparameters Selection}

Based on the validation and test sets we have split, we tune the hyperparameter, $\alpha$, on the validation set. We adjust it in intervals of 0.05 in absolute value and select the value of $\alpha$ that performs best on the validation set to apply to the test set.

% \section{Example Outputs}
% \label{appendix: example outputs}

All the experiments in this work were conducted on four 80GB NVIDIA A800 GPUs.

\end{document}